\documentclass{article}
\usepackage[nocompress]{cite}

\usepackage{times}
\usepackage{url}
\usepackage{latexsym}

\usepackage{amssymb} 
\usepackage{amsmath} 
\usepackage{amsthm}

\usepackage{array}
\usepackage{color}
\usepackage{multirow}

\usepackage{algorithmicx}
\usepackage{algpseudocode}

\usepackage{graphicx}
\graphicspath{{fig/}}

\usepackage{multirow}
\newcommand{\equref}{\eqref}

\usepackage{spconf,amsmath,graphicx}

\title{Learning neural trans-dimensional random field language models with noise-contrastive estimation}
\name{Bin Wang, Zhijian Ou\thanks{This work is supported by NSFC grant 61473168.}}
\address{Speech Processing and Machine Intelligence (SPMI) Lab, Tsinghua University, Beijing, China. \\
         wangbin12@mails.tsinghua.edu.cn, ozj@tsinghua.edu.cn}
\begin{document}
\ninept
\maketitle
\begin{abstract}
Trans-dimensional random field language models (TRF LMs) where sentences are modeled as a collection of random fields,
have shown close performance with LSTM LMs in speech recognition and are computationally more efficient in inference.
However, the training efficiency of neural TRF LMs is not satisfactory, which limits the scalability of TRF LMs on large training corpus.
In this paper, several techniques on both model formulation and parameter estimation are proposed to improve the training efficiency and the performance of neural TRF LMs.
First, TRFs are reformulated in the form of exponential tilting of a reference distribution.
Second, noise-contrastive estimation (NCE) is introduced to jointly estimate the model parameters and normalization constants.
Third, we extend the neural TRF LMs by marrying the deep convolutional neural network (CNN) and the bidirectional LSTM into the potential function to extract the deep hierarchical features and bidirectionally sequential features.
Utilizing all the above techniques enables the successful and efficient training of neural TRF LMs on a 40x larger training set with only 1/3 training time
and further reduces the WER with relative reduction of 4.7\% on top of a strong LSTM LM baseline.
\end{abstract}

\section{Introduction}
\label{sec:intro}

Statistical language models are a crucial component in many applications, such as automatic speech recognition (ASR) and machine translation (MT), by encoding the linguistic regularities in terms of the joint probability of words in a sentence.
Currently, the recurrent neural network approach, which follows the directed graphical modeling approach,
has shown significant perplexity reductions over the classic n-gram LMs in various benchmarks such as
the Penn Tree Bank \cite{lstmdropout} and the Google One Billion corpus \cite{jozefowicz2016exploring},
and also achieves the state-of-the-art word error rates (WERs) in ASR \cite{kurata2017empirical}.

In contrast, a new trans-dimensional random field (TRF) LM \cite{Bin2015,Bin2017,Bin2017ASRU} has been proposed in the undirected graphical modeling approach,
where sentences are modeled as a collection of random fields
and the joint probability is defined in terms of potential functions. 
With the power of flexibly supporting rich features,
neural TRF LMs\cite{Bin2017ASRU} with a nonlinear potential function defined by a deep convolutional neural network (CNN),
outperform the n-gram LMs significantly
and perform close to LSTM LMs and are computational more efficient in inference.

However the training speed of neural TRF LMs is not satisfactory, which is caused by several factors.
First, the estimation of TRFs depends on the newly developed augmented stochastic approximation (AugSA) algorithm \cite{Bin2017},
where the samples generated based on the Markov chain Monte Carlo (MCMC) theory
are used to update the parameters.
The convergence of such approach heavily depends on the quality and quantity of the samples and
usually needs a lot of training iterations.
Second, the trans-dimensional mixture sampling method (TransMS) which is used to generate samples of varying dimensions
is computational expensive,
even when the joint stochastic approximation (JSA) \cite{xu2016joint} strategy is introduced to improve its efficiency.
This is partly due to that the sampling operation depends on the current model parameters,
and can not be performed beforehand or be parallelized with other training operations.
Third, learning the neural TRF LMs need to optimize a non-convex objective function,
which is much harder than learning the discrete TRF LMs which is a convex optimization.
All the above factors limit the training efficiency of neural TRFs as well as their scalability on large training corpus.

In this paper, the following contributions are made to improve both the training efficiency and the performance of neural TRF LMs.
First, TRFs are defined in the form of exponential tilting of a reference distribution.
As a result, only the difference between the data distribution and the reference distribution needs to be fitted by TRFs,
which is much simpler than fitting the empirical distribution directly.
Second, noise-contrastive estimation (NCE) \cite{nce} is introduced to train TRF LMs
by optimizing a discriminator between the real sentences drawn from the training set and the noise sentences drawn from a noise distribution.
The noise distribution in NCE is independent of the model distribution and
hence drawing noise sentences can be parallelized with model estimation to accelerate the training process significantly.
Meanwhile, the normalization constants in TRFs can be treated as the normal parameters and be jointly optimized with the model parameters during NCE training.
Third, we enhance the neural TRF in \cite{Bin2017ASRU} by marrying deep CNN and  bidirectional LSTM into the potential function to extract the deep hierarchical features and bidirectionally sequential features.

Two experiments are designed in the paper,
including a pilot experiment called short-word morphology to validate the NCE in TRF situation and
an ASR experiment on CHiME-4 Challenge data to evaluate the scalability of neural TRF LMs.
First, the pilot experiment reveals the effectiveness of NCE in TRF training where more than one normalization constants need to be estimated simultaneously.
Then in ASR,
a neural TRF LM with a LSTM LM as the reference distribution is trained using NCE in 40x larger training corpus with only a third of training time, compared with the results reported in \cite{Bin2017ASRU}.
The results show that the neural TRF models and the directed graphical modeling models are complementary. The lowest WERs are achieved by combining the neural TRF LMs with n-gram LMs and LSTM LMs,
with relative reduction of 4.7\% over state-of-the-art LSTM LMs, i.e. combining the n-gram LMs and LSTM LMs.

The rest of the paper is organized as follows.
We first discuss relate work in Section \ref{sec:RelatedWork}.
Then in Section \ref{sec:ModelDefination}, we describe the new formulation of neural TRFs with the neural network potential.
The NCE training method is introduced in Section \ref{sec:nce}.
After presenting the experimental results in Section \ref{sec:Experiments},
the conclusions are made in Section \ref{sec:Conclusion}.

\section{Related work}
\label{sec:RelatedWork}

The noise-contrastive estimation is first proposed in \cite{nce} for the estimation of  unnormalized statistical models, whose normalization constants can not be obtained in closed form.
In language modeling, NCE is used to train the conditional neural network (NN) LMs, such as the feedforward neural network LMs \cite{vaswani2013decoding} and LSTM LMs \cite{zoph2016simple},
by treating the learning as a binary classification problem between the target words and the noise samples.
As the normalization terms of the prediction probabilities are dependent on the context and can not be enumerated,
an approximate approach is to freeze them to an empirical value, which is 1 in \cite{zoph2016simple} and $e^9$ in \cite{chen2015recurrent}.
Moreover,
\cite{zoph2016simple} accelerates the NCE training by sharing the noise samples for each target word in the mini-batch to reduce the sample times.
Then \cite{oualil2017batch} further omits the sampling process by taking the other target words in current mini-batch as the noise samples.
In this paper, NCE is evaluated in the estimation of TRF LMs, which are defined on the trans-dimensional state space.
To our best knowledge, this is the first time that NCE is applied in the trans-dimensional setting.

The idea of defining a model in the form of exponential tilting of a reference distribution
has been studied in natural image generative modeling \cite{dai2014generative,xie2016theory} where the reference distribution is set to the Gaussian white noise distribution.
Similar formulations have been used in language modeling with a class n-gram LM as the reference,
including the maximum entropy LMs \cite{berger1996maximum} and the whole-sentence LMs \cite{rosenfeld1997whole,amaya2001improvement,ruokolainen2010using}.
In this paper, a LSTM LM is used as the reference distribution to define a neural TRF LM.

\section{Neural trans-dimensional random field LMs}
\label{sec:ModelDefination}

Different form the neural TRFs in \cite{Bin2017ASRU}, here the joint probability of a sequence $x^l$ and its length $l$ ($l=1,\cdots,m$) is assumed to be in the form of exponential tilting of a reference distribution $q(x^l)$:
    \begin{equation} \label{eq:model_all}
    p(l, x^l;\theta) = \frac{\pi_l}{Z_l(\theta)} q(x^l) e^{\phi(x^l;\theta)}
    \end{equation}
where
$x^l = (x_1, \cdots, x_l)$ is a word sequence of length $l$,
$\pi_l$ is the prior length probability,
$\theta$ indicates the set of parameters,
$\phi$ is the potential function,
and $Z_l(\theta)$ is the normalization constant of length $l$, i.e. $Z_l(\theta) = \sum_{x^l} q(x^l) e^{\phi(x^l; \theta)}$.
Different with the TRFs defined in \cite{Bin2017,Bin2017ASRU}, a reference distribution $q(x^l)$ is introduced as the baseline distribution,
and the role of TRFs is to fit the difference between the data distribution and the reference distribution.
If $q(x^l)$ is a good approximation of the data distribution, such as the LSTM LMs used in our experiments,
fitting the difference between the data distribution and the reference distribution $q(x^l)$ shall be much simpler than fitting the data distribution directly.

\begin{figure}[t]
    \centering
    \includegraphics[width=0.85\linewidth]{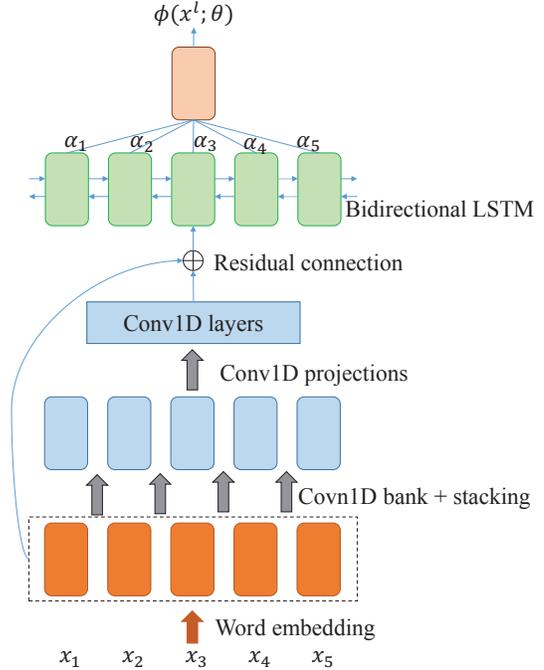}
    \caption{Neural network architecture used to define the potential function $\phi(x^l;\theta)$}
    \label{fig:nn}
\end{figure}

To compute the potential function, we define a neural network by combining the deep CNN structure and the bidirectional LSTM structure to extract both the deep hierarchical features and bidirectionally sequential features, as shown in Figure \ref{fig:nn}.
The architecture is motivated by the encoder module in \cite{wang2017tacotron} with some simplifications and is described as follows.

In the bottom, a deep CNN is used to extract deep hierarchical features,
whose architecture is similar to the neural network used in \cite{Bin2017ASRU} except that the weighted summation in the ``CNN-stack'' module is removed.
There are three steps.
First, each word $x_i$ ($i=1, \cdots, l$) in a sentence is mapped to an embedding vector.
Then, these embedding vectors in a sentence are fed into a ``CNN-bank'' module, which contains a set of 1-D convolutional filters with widths ranging from $1$ to $K$.
These filters explicitly model local contextual information (akin to modeling unigrams, bigrams, up to K-grams) \cite{kim2016character}.
Third, the output feature maps from multiple filters with varying widths are spliced together,
and fed into a few fixed-width 1-D convolutions to further extract hierarchical features, which resembles the ``CNN-stack'' module in \cite{Bin2017ASRU}.
The output of the last 1-D convolution is added with the word embedding vectors via a residual connection.

A bidirectional LSTM (BLSTM) is stacked on top of the deep CNN to extract long-range sequential features from the forward and backward contexts.
Note that all the convolutions mentioned above are \emph{half convolutions}
\footnote{http://deeplearning.net/software/theano/library/tensor/nnet/conv.html}, i.e. convolutions are performed after padding zeros to the beginning and the end of the input sequences to preserve the time resolution.

Finally, the attention mechanism is introduced to summate the feature vectors of BLSTM at all positions,
and the potential function $\phi(x^l;\theta)$ is computed as follows.
    \begin{align}
    \phi(x^l; \theta) &= \lambda^T \sum_{i=1}^l \alpha_i h[:,i]  + c \\
    \alpha_i &= \beta^T h[:,i], \quad i=1,\cdots,l
    \end{align}
where $h \in R^{2d \times l}$ is the hidden vectors in the BLSTM, $d$ is the number of hidden units of the BLSTM,
$h[:,i]$ is the $i$-th column of $h$ and
$\lambda, \beta \in R^{2d}$, $c \in R$ are the parameters.
In summary, $\theta$ denotes the collection of all the parameters defined in the neural networks.

\section{Noise-Contrastive Estimation}
\label{sec:nce}

Noise-contrastive estimation (NCE) is proposed by \cite{nce} for the estimation of unnormalized statistical models,
and has been successfully used in the estimation of neural network LMs \cite{vaswani2013decoding, oualil2017batch, zoph2016simple, chen2015recurrent}.
The basic idea of NCE is ``learning by comparison'', i.e. to perform nonlinear logistic regression to discriminate between the data samples drawn from the training set
and the noise samples drawn from a known noise distribution.
The normalization constants can be treated as the normal parameters and updated together with the model parameters.

To apply NCE to estimate the neural TRFs defined in \equref{eq:model_all},
first we rewrite the formulation in \equref{eq:model_all} by introducing a new parameter $\zeta$:
    \begin{equation} \label{eq:model_all_zeta}
    p(l,x^l;\theta,\zeta) = \pi_l q(x^l) e^{\phi(x^l,;\theta)-\zeta_l}
    \end{equation}
where $\zeta=(\zeta_1, \cdots, \zeta_l)$ denotes the hypothesized values of the true logarithmic normalization constants $\zeta_l^* = \log Z_l$,
which can be estimated in NCE.
Assuming for each sequence in the training set, $\nu$ noise sequences of varying lengths are generated from a noise distribution $p_n(l, x^l)$.
Then the probabilities of a sequence $(l,x^l)$ belonging to the data distribution $P(C=0|l, x^l; \theta, \zeta)$ and the noise distribution $P(C=1|l, x^l; \theta, \zeta)$ are given by
    \begin{align}
    P(C=0|l, x^l; \theta, \zeta) &= \frac{p(l, x^l; \theta, \zeta)}{p(l, x^l; \theta, \zeta) + \nu p_n(l, x^l)} \\
    P(C=1|l, x^l; \theta, \zeta) &= 1 - P(C=0|l, x^l; \theta, \zeta)
    \end{align}

Given the training set $D$, NCE maximizes the following conditional log-likelihood:
    \begin{equation}
    \begin{split}
    J(\theta,\zeta) = \frac{1}{|D|} \sum_{(l,x^l) \in D} \log P(C=0|l,x^l;\theta,\zeta) + \\
                  \nu \frac{1}{|B|} \sum_{(l,x^l) \in B} \log P(C=1|l,x^l;\theta,\zeta)
    \end{split}
    \end{equation}
where $B$ is the noise sample set, which is generated from the noise distribution $p_n(l, x^l)$,
$|D|$ and $|B|$ are the number of samples in $D$ and $B$ respectively, and satisfy $\nu = |B|/|D|$.
To maximize the objective $J(\theta,\zeta)$, the gradient with respect to $\theta$ and $\zeta$ can be computed as follows:
    \begin{equation}
    \begin{split}
    \frac{\partial J(\theta,\zeta)}{\partial \theta} =
                    \sum_{(l,x^l) \in D} w_t(l,x^l;\theta,\zeta) \frac{\partial \phi(x^l; \theta)}{\partial \theta} + \\
                    \sum_{(l,x^l) \in B} w_n(l,x^l;\theta,\zeta) \frac{\partial \phi(x^l; \theta)}{\partial \theta}
    \end{split}
    \end{equation}
    \begin{equation}
    \begin{split}
    \frac{\partial J(\theta,\zeta))}{\partial \zeta_j} =
                    -\sum_{(l,x^l) \in D } w_t(l,x^l;\theta,\zeta) 1(l=j) - \\
                    \sum_{(l,x^l) \in B} w_n(l,x^l;\theta,\zeta)  1(l=j)
    \end{split}
    \end{equation}
where
    \begin{equation}
    \left \{
        \begin{split}
        w_t(l,x^l;\theta,\zeta) &= \frac{1-P(C=0|l,x^l;\theta,\zeta)}{|D|} \\
        w_n(l,x^l;\theta,\zeta) &= -\frac{P(C=0|l,x^l;\theta,\zeta)}{|D|}
        \end{split}
    \right.
    \end{equation}
and $1(l=j)$ is 1 if $l=j$ and 0 otherwise.
The gradient of the potential function $\phi(x^l;\theta)$ with respect to the parameters $\theta$ can be computed through the back-propagation algorithm.
Then any gradient method can be used to optimize the parameters and normalization constants, such as stochastic gradient descent (SGD) or Adam \cite{adam}.

In our experiments, the noise distribution is defined as
    \begin{equation} \label{eq:noise}
    p_n(l, x^l) = \pi_l p_n(x^l),
    \end{equation}
where $\pi_l, l=1, ..., m$ is the prior length probability and $p_n(x^l)$ is a n-gram LM.
This makes the noise distribution $p_n(l, x^l)$ close to the data distribution,
and we can achieve a good performance with a medium sample number $\nu$, such as 20 in our following experiments,
as suggested in \cite{nce}.

\section{Experiments}
\label{sec:Experiments}

In this section, two experiments are conducted to evaluate NCE training for neural TRFs.
First, a pilot experiment - short-word morphology is designed to exactly evaluate the performance of NCE.
Then neural TRFs and NCE training are applied to language modeling in ASR using CHiME-4 Challenge data \cite{chime4}.

\subsection{Short-word morphology}

In this section, we design a simulation experiment to validate the NCE training for neural TRF models,
where more than one normalization constants need to be estimated simultaneously.
The training set and valid set include 5,143 and 69 different English words with at most 3 characters respectively, which are extracted from English Gigaword dataset \cite{gigaword}.
Every word is decomposed to a character sequence and the objective of this experiment is to assign a probability to the character sequences.
The vocabulary contains a total of 28 symbols, i.e. the 26 English letters and two auxiliary symbols - the beginning symbol and the end symbol, which are added to the beginning and the end of every character sequence respectively.
As the length of each sequences is no more than 3, the normalization constants $Z_l, l=1, 2, 3$ can be calculate exactly.

\begin{figure}[t]
\begin{minipage}[b]{.5\linewidth}
  \centering
  \centerline{\includegraphics[width=1.1\linewidth]{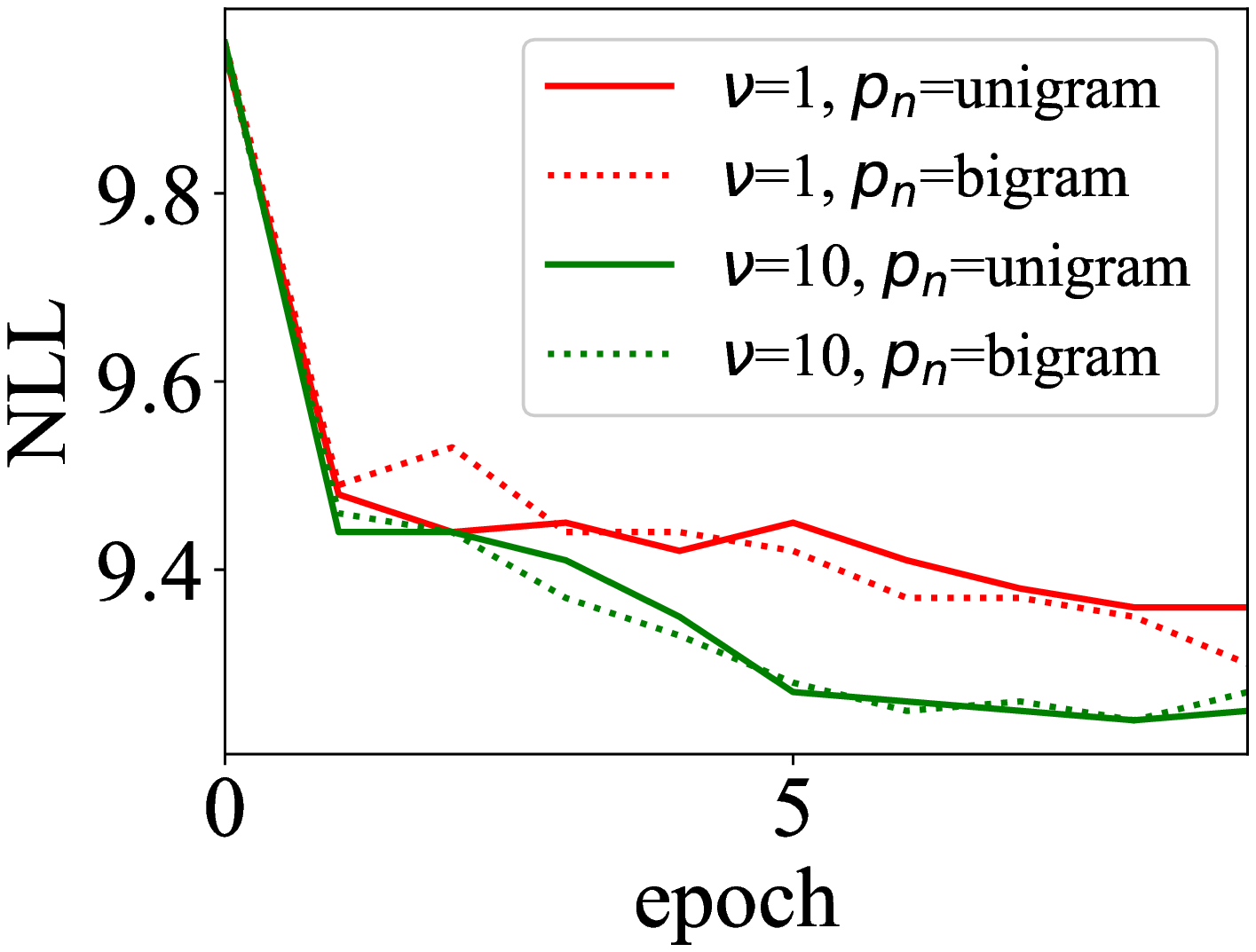}}
  \centerline{(a)}
\end{minipage}
\begin{minipage}[b]{.5\linewidth}
  \centering
  \centerline{\includegraphics[width=1.1\linewidth]{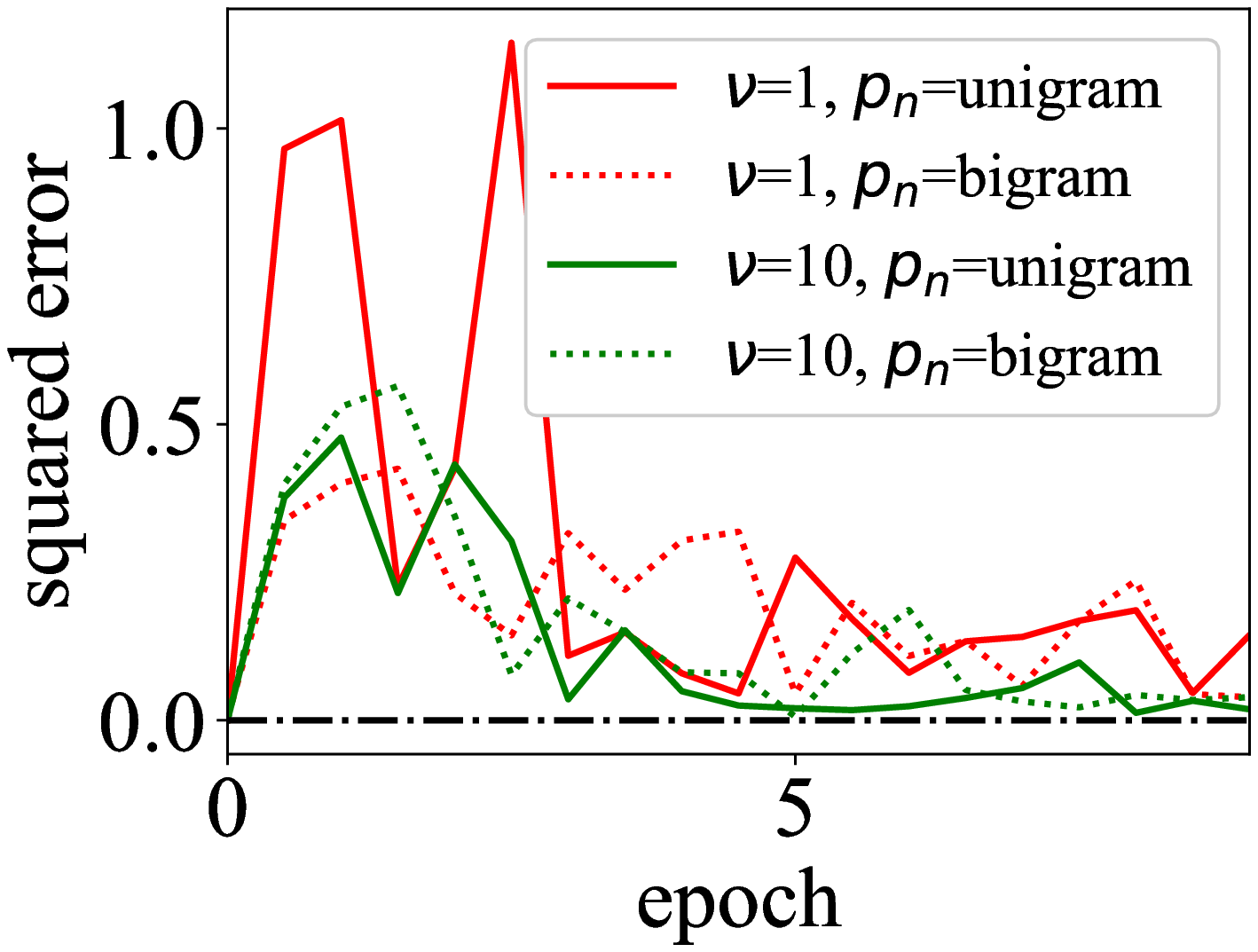}}
  \centerline{(b)}
\end{minipage}
\caption{The results of short-word morphology.
        (a) is the negative log-likelihood (NLL) on the valid set using the true normalization constants.
        (b) is the squared error between the estimated normalization constants $\zeta$ and the true normalization constants $\zeta^*$.
        Different configurations for number of noise samples $\nu=1,10$ and noise distribution $p_n(x^l)=unigram,bigram$ are investigated.
        }
\label{fig:exp1}
\end{figure}

A neural TRF defined in section \ref{sec:ModelDefination} is applied to model these character sequences, with $q(x^l)$ being uniform.
For the potential function,
each character is embedded to a 16 dimensional vector and directly fed into the BLSTM without passing the CNN layers.
The BLSTM contains 1 hidden layers for each direction and $16$ hidden units for each layer.
The prior length distribution $\pi_l$ in model distribution \eqref{eq:model_all_zeta} and the noise distribution \eqref{eq:noise} are both set to the empirical length distribution.
The parameters $\theta$ are initialized randomly within an interval from -0.1 to 0.1,
and the initial value of normalization constants are $\zeta_l = l \times \log V$ where $V=28$ denotes the vocabulary size.
We use Adam to update the parameter $\theta$ and the normalization constants $\zeta$ with the mini-batch size being 10.
The learning rates for $\theta$ and $\zeta$ are fixed to $0.001$ and $0.01$ respectively.
We investigate different configurations for sample number $\nu$ and the noise distribution $p_n(x^l)$.
The results are summarized in Figure \ref{fig:exp1}.
The main conclusions are as follows.

First, NCE performs well in TRF training, with both the NLL and normalization constants converging after several training epochs.
Second, increasing the sample number $\nu$ form 1 to 10 can lead to a lower NLL (Figure \ref{fig:exp1}(a))
and make the normalization constants $\zeta$ converge fast to the true $\zeta^*$ (Figure \ref{fig:exp1}(b)).
Third, for a small sample number $\nu=1$,
changing the noise distribution $p_n(x^l)$ from unigram (red solid line) to bigram (red dot line) will improve the convergence of NCE.
If we increase the sample number to $\nu=10$,
the choice of the noise distribution makes less differences.
The above observations of NEC in TRF training is consistent with \cite{nce}

\subsection{Neural TRF LMs in speech recognition}

In this section,
we evaluate the performance and scalability of neural TRF LMs trained by NCE over CHiME-4 Challenge data.
The training corpus for language modeling contains about 37 millions tokens,
which is about 40 times of the Penn Treebank (PTB) training set used in \cite{Bin2017ASRU}.
The vocabulary is limited to 5 K words, including a special token $\langle$UNK$\rangle$ denoting the out-of-vocabulary words.

For evaluation in terms of speech recognition WERs,
various LMs are applied to rescore the 100-best lists from recognizing CHiME-4 development and test data.
For each utterance, the 100-best list of candidate sentences are generated by the multi-channel ASR system developed by our team under CHiME-4 Challenge rule, which is detailed in \cite{xiangthu}.
All the hyper-parameters of training LMs, such as learning rates and training epochs, are tuned on the development utterances to achieve the lowest WER.
In CHiME-4 Challenge, different systems are compared in terms of the WERs on the real test set, which consists of speech recordings collected in real environments.

\begin{table}[t]
    \centering
    \begin{tabular}{c|c|p{0.5\linewidth}}
        \hline
        $\pi_l$  & \multicolumn{2}{l}{empirical length probabilities} \\
        \hline
        \multirow{2}{*}{$q(x^l)$}  &  \multicolumn{2}{p{0.8\linewidth}}{a LSTM with 512 embedding size, 2 hidden layers and 512 hidden units per layer \cite{lstmdropout}} \\
        \hline
        \multirow{8}{*}{$\phi(x^l;\theta)$} & embedding size  & 200 \\
        \cline{2-3}
        & \multirow{2}{*}{Conv1D bank} & cnn-$k$-128-ReLU, \newline with $k$ ranging from 1 to 10 \\
        \cline{2-3}
        & \multirow{3}{*}{Conv1D layers} & cnn-3-128-ReLU $\rightarrow$ \newline cnn-3-128-ReLU $\rightarrow$ \newline cnn-3-128-ReLU \\
        \cline{2-3}
        & \multirow{1}{*}{BLSTM} & 1 layer and 128 hidden units \\
        \hline
    \end{tabular}
    \caption{
    The configuration of neural TRFs.
    ``cnn-$k$-$n$-ReLU'' denotes a 1-D CNN with filter width $k$, output dimension $n$ and rectified linear unit (ReLU) activation.
    ``$A \rightarrow B$'' denotes that the output of layer $A$ is fed into layer $B$.
    }
    \label{tab:config}
\end{table}

\begin{table}[t]
    \centering
    \begin{tabular}{l|c|c|c|c}
        \hline
        \multirow{2}{*}{model} & \multicolumn{2}{c|}{Dev} & \multicolumn{2}{c}{Test} \\
        \cline{2-5}
                    & real & simu & real & simu \\
        \hline
        KN5          & 5.03 & 4.79  & 7.38  & 5.78 \\
        LSTM2x512    & 3.63 & 3.24  & 5.70  & 4.53 \\
        neural TRF   & 3.53 & 3.20  & 5.68  & 4.36 \\
        KN5+LSTM2x512 & 3.56 &   3.29 &   \textbf{5.71} &   4.18 \\
        KN5+neural TRF & 3.53 & 3.22 & 5.54  & 4.20 \\
        KN5+LSTM2x512+neural TRF & 3.42 &   3.10 &   \textbf{5.44} &   4.13 \\
        \hline
    \end{tabular}
    \caption{
        Speech recognition WERs on CHiME-4 Challenge data.
        ``Dev'' denotes the development set and ``Test'' denotes the test set.
        ``real'' denotes the speech recorded in real environments and ``simu'' denotes the simulated speech.
    }
    \label{tab:wer}
\end{table}

The configuration of the neural TRF LMs used in this experiments are listed in Table \ref{tab:config}.
First a LSTM LM \cite{lstmdropout} with 2 hidden layers and 512 hidden units per layer is trained on the training set using SGD method,
and serves as the reference distribution $q(x^l)$ of neural TRF LMs in \equref{eq:model_all}.
Then NCE with $\nu=20$ is used to train the neural TRF LM by fixing the parameters of the reference distribution $q(x^l)$,
and updating the model parameters $\theta$ and normalization constants $\zeta$ simultaneously based SGD method.
The learning rates for both $\theta$ and $\zeta$ are same
which are initialized to 0.01 and reduce by half per epoch.
The parameters $\theta$ are initialized randomly within an interval from -0.1 to 0.1,
and the normalization constants are initialized to $\zeta_l = l$ for $l=1,\cdots,m$.
We perform the training process for 2 epoches before the neural TRF achieves the lowest WER in the development set.
The total training time is 1 day, which is one third of the training time reported in \cite{Bin2017ASRU}.

The WERs of various LMs are shown in Table \ref{tab:wer},
including a 5-gram LM with modified Kneser-Ney smoothing \cite{chen1999empirical} (denoted by ``KN5''),
and the reference LSTM LMs $q(x^l)$ (denoted by ``LSTM2x512'' to emphasize the hidden layers and hidden units).
First, the LSTM LM achieves significant WER reduction compared with ``KN5'' with relative WER reduction 22.8\% on real test set.
Combining the LSTM LM and the n-gram LM leads to the state-of-the-art results, denoted by ``KN5+LSTM2x512''.
Building on top of the LSTM LM,
our neural TRF LM can further reduce the WER with only 2-epoch NCE training.
Remarkably,
the lowest WER 5.44\% on real test set is achieved by combining neural TRF LMs with n-gram LMs and LSTM LMs (denoted by ``KN5+LSTM2x512+neural TRF''),
with relative reduction of 4.7\% compared to the strong state-of-the-art system ``KN5+LSTM2x512''.
This reveals that neural TRF LMs and various LMs in the directed graphical modeling approach are complementary, and a combination of them leads to WER reduction.

\section{Conclusion}
\label{sec:Conclusion}

This paper presents our continuous effort to develop the TRF approach to language modeling. We make the following contributions to improve the convergence and scalability of neural TRF LMs.
First, a reference distribution is introduced to serve as a baseline distribution.
Then NCE is used to estimate the parameters and normalization constants jointly.
Finally, new neural network structures are investigated in neural TRF LMs, by combing the CNN and bidirectional LSTM into the potential function.
Utilizing all these techniques leads to the superior performance of TRF LMs in the CHiME-4 Challenge dataset, which is a supporting evidence of the power of TRF LMs on a medium size training set.

It is worthwhile to further investigate techniques to improve the training efficiency of TRF LMs.
Moreover, integrating richer nonlinear and structured features is an important future direction.
The neural TRF models can also be applied to other sequential
and trans-dimensional data modeling tasks in general.

\bibliographystyle{IEEEbib}
\bibliography{RF}

\begin{thebibliography}{10}

\bibitem{lstmdropout}
Wojciech Zaremba, Ilya Sutskever, and Oriol Vinyals,
\newblock ``Recurrent neural network regularization,''
\newblock {\em arXiv preprint arXiv:1409.2329}, 2014.

\bibitem{jozefowicz2016exploring}
Rafal Jozefowicz, Oriol Vinyals, Mike Schuster, Noam Shazeer, and Yonghui Wu,
\newblock ``Exploring the limits of language modeling,''
\newblock {\em arXiv preprint arXiv:1602.02410}, 2016.

\bibitem{kurata2017empirical}
Gakuto Kurata, Abhinav Sethy, Bhuvana Ramabhadran, and George Saon,
\newblock ``Empirical exploration of novel architectures and objectives for
  language models,''
\newblock in {\em Proc. INTERSPEECH}, 2017.

\bibitem{Bin2015}
Bin Wang, Zhijian Ou, and Zhiqiang Tan,
\newblock ``Trans-dimensional random fields for language modeling,''
\newblock in {\em Proc. Annu. Meeting of the Association for Computational
  Linguistics (ACL)}, 2015.

\bibitem{Bin2017}
Bin Wang, Zhijian Ou, and Zhiqiang Tan,
\newblock ``Learning trans-dimensional random fields with applications to
  language modeling,''
\newblock {\em IEEE Transactions on Pattern Analysis and Machine Intelligence
  (PAMI)}, 2017.

\bibitem{Bin2017ASRU}
Bin Wang and Zhijian Ou,
\newblock ``Language modeling with neural trans-dimensional random fields,''
\newblock in {\em IEEE Automatic Speech Recognition and Understanding
  Workshop}, 2017.

\bibitem{xu2016joint}
Haotian Xu and Zhijian Ou,
\newblock ``Joint stochastic approximation learning of helmholtz machines,''
\newblock {\em International Conference on Learning Representations (ICLR)
  Workshop Track}, 2016.

\bibitem{nce}
Michael Gutmann and Aapo Hyv{\"a}rinen,
\newblock ``Noise-contrastive estimation: A new estimation principle for
  unnormalized statistical models,''
\newblock in {\em Proceedings of the Thirteenth International Conference on
  Artificial Intelligence and Statistics}, 2010.

\bibitem{vaswani2013decoding}
Ashish Vaswani, Yinggong Zhao, Victoria Fossum, and David Chiang,
\newblock ``Decoding with large-scale neural language models improves
  translation.,''
\newblock in {\em EMNLP}, 2013.

\bibitem{zoph2016simple}
Barret Zoph, Ashish Vaswani, Jonathan May, and Kevin Knight,
\newblock ``Simple, fast noise-contrastive estimation for large rnn
  vocabularies.,''
\newblock in {\em HLT-NAACL}, 2016.

\bibitem{chen2015recurrent}
Xie Chen, Xunying Liu, Mark~JF Gales, and Philip~C Woodland,
\newblock ``Recurrent neural network language model training with noise
  contrastive estimation for speech recognition,''
\newblock in {\em Acoustics, Speech and Signal Processing (ICASSP), 2015 IEEE
  International Conference on}, 2015.

\bibitem{oualil2017batch}
Youssef Oualil and Dietrich Klakow,
\newblock ``A batch noise contrastive estimation approach for training large
  vocabulary language models,''
\newblock {\em arXiv preprint arXiv:1708.05997}, 2017.

\bibitem{dai2014generative}
Jifeng Dai, Yang Lu, and Ying-Nian Wu,
\newblock ``Generative modeling of convolutional neural networks,''
\newblock {\em arXiv preprint arXiv:1412.6296}, 2014.

\bibitem{xie2016theory}
Jianwen Xie, Yang Lu, Song-Chun Zhu, and Yingnian Wu,
\newblock ``A theory of generative convnet,''
\newblock in {\em International Conference on Machine Learning}, 2016.

\bibitem{berger1996maximum}
Adam~L Berger, Vincent J~Della Pietra, and Stephen A~Della Pietra,
\newblock ``A maximum entropy approach to natural language processing,''
\newblock {\em Computational linguistics}, vol. 22, no. 1, pp. 39--71, 1996.

\bibitem{rosenfeld1997whole}
Ronald Rosenfeld,
\newblock ``A whole sentence maximum entropy language model,''
\newblock in {\em Proc. Automatic Speech Recognition and Understanding (ASRU)},
  1997.

\bibitem{amaya2001improvement}
Fredy Amaya and Jos{\'e}~Miguel Bened{\'\i},
\newblock ``Improvement of a whole sentence maximum entropy language model
  using grammatical features,''
\newblock in {\em Proc. Ann. Meeting of the Association for Computational
  Linguistics (ACL)}, 2001.

\bibitem{ruokolainen2010using}
Teemu Ruokolainen, Tanel Alum{\"a}e, and Marcus Dobrinkat,
\newblock ``Using dependency grammar features in whole sentence maximum entropy
  language model for speech recognition.,''
\newblock in {\em Baltic HLT}, 2010.

\bibitem{wang2017tacotron}
Yuxuan Wang, RJ~Skerry-Ryan, Daisy Stanton, Yonghui Wu, Ron~J Weiss, Navdeep
  Jaitly, Zongheng Yang, Ying Xiao, Zhifeng Chen, Samy Bengio, et~al.,
\newblock ``Tacotron: A fully end-to-end text-to-speech synthesis model,''
\newblock {\em arXiv preprint arXiv:1703.10135}, 2017.

\bibitem{kim2016character}
Yoon Kim, Yacine Jernite, David Sontag, and Alexander~M Rush,
\newblock ``Character-aware neural language models,''
\newblock in {\em Thirtieth AAAI Conference on Artificial Intelligence}, 2016.

\bibitem{adam}
Diederik Kingma and Jimmy Ba,
\newblock ``Adam: A method for stochastic optimization,''
\newblock {\em arXiv:1412.6980 [cs.LG]}, 2014.

\bibitem{chime4}
``Chime-4 speech separation and recognition challenge,''
  \url{http://spandh.dcs.shef.ac.uk/chime_challenge/chime2016/index.html}.

\bibitem{gigaword}
``Ldc english gigaword,'' \url{https://catalog.ldc.upenn.edu/LDC2003T05}.

\bibitem{xiangthu}
Hongyu Xiang, Bin Wang, and Zhijian Ou,
\newblock ``The thu-spmi chime-4 system: Lightweight design with advanced
  multi-channel processing, feature enhancement, and language modeling,''
\newblock in {\em CHiME-4 Workshop}, 2016.

\bibitem{chen1999empirical}
Stanley~F. Chen and Joshua Goodman,
\newblock ``An empirical study of smoothing techniques for language modeling,''
\newblock {\em Computer Speech \& Language}, vol. 13, pp. 359--394, 1999.

\end{thebibliography}

\end{document}